# Image Inpainting by Multiscale Spline Interpolation

Ghazale Ghorbanzade, Zahra Nabizadeh, Nader Karimi, Shadrokh Samavi

Isfahan University of Technology

Isfahan, 84156-83111 Iran

*Abstract—* **Recovering the missing regions of an image is a task that is called image inpainting. Depending on the shape of missing areas, different methods are presented in the literature. One of the challenges of this problem is extracting features that lead to better results. Experimental results show that both global and local features are useful for this purpose. In this paper, we propose a multi-scale image inpainting method that utilizes both local and global features. The first step of this method is to determine how many scales we need to use, which depends on the width of the lines in the map of the missing region. Then we apply adaptive image inpainting to the damaged areas of the image, and the lost pixels are predicted. Each scale is inpainted and the result is resized to the original size. Then a voting process produces the final result. The proposed method is tested on damaged images with scratches and creases. The metric that we use to evaluate our approach is PSNR. On average, we achieved 1.2 dB improvement over some existing inpainting approaches.**

*Keywords— multi-scale, image inpainting, interpolation, mask, missing pixels*

## I. INTRODUCTION

Due to the increasing use of digital images instead of printed ones, image enhancement techniques have received a lot of attention. Among a variety of image enhancement techniques, image inpainting or completion could be used to predict the lost pixels in an image. In this algorithm, there is a lost region $\Omega$ in image $\mu$. If the pixels in this region known as $p(i,j)$, then the purpose is to fill these pixels with the source pixels $\Omega^c$ which $\Omega^c \cap \Omega = \mu$. The best algorithm is the one that utilizes the best combination of intensity value and gradient information of pixels in $\Omega^c$ to predict the value of $p(i,j)$ in $\Omega$. According to this definition, there are different types of degradation for $\Omega$ in images. These degradations include (i) scratches and crease, (ii) missing blocks, and (iii) text and small blanks [1]. These categories of missing pixels originate from their applications. There are different types of applications for each category, for example, enhancement of old and historical images, generating new images by adding or removing objects, and erasing texts or watermarks from images. For each application, various algorithmic and learning methods are proposed. In the following, based on the degradation type, some of the existing methods will be described.

### A. Scratch Removing Methods

The shape of lost pixels in scratches is usually a line, a curve, or a point. In other words, scratches are small and narrow regions. For removing it, different algorithmic methods are proposed. In some frames of video, scratch lines are a common defect. These damages occur as a result of the transport of the film or the developing process [2]. These scratches are vertical lines with a width of 3 to 10 pixels. In [2], Hongying et al. proposed a method that could detect the lost pixels by highlighting the scratch regions, and after that, by using the p-Laplace operator, lost pixels are predicted. In some images, an object is selected to remove for example omitting eyeglasses from the frontal face image. In [3], a method is presented to detect eyeglasses and remove them from images. In this method, the eye region detector is used to locate the eye and glasses. Then, a technique based on Markov-chain Monte Carlo is used to accurately find the points that are to be replaced. Finally, by statistical analysis, a mapping of pixels, between the face image with the glass to the image without the glass, is learned. In [4], by using one and two-dimensional cubic splines, the interpolation of lost pixels is done. In this work, fixed neighbors are considered for lost pixels. Akbari et al. [5] proposed a method that selects a correlated patch for the lost pixels.

### B. Block Removing Based Methods

In some applications, for example, adding or removing objects, a block of pixels is lost. In this situation, it needs information about the whole image, scene, texture, and other features to predict the lost block. Different algorithmic and learning methods are proposed to address this problem. But recently, due to the development of deep neural networks (DNN), the use of these has received more attention in many applications, especially image inpainting. In different layers of DNN, features with various levels of details are extracted. In recent approaches based on DNN, background information, which includes texture information, is used for predicting missing regions. In these approaches, the local patch information is used for missing blocks, but in some situations, lost zones and background information are not similar. In [6], a multi-scale image contextual attention learning (MUSICAL) is proposed to address this problem. In MUSICAL, besides local features, global features are also extracted from multi-scales of the input image. So, by using local and global information, the prediction of missing blocks is improved [6]. Various types of DNNs are useful for image inpainting. The most applicable one is the generative adversarial network (GAN). This network consists of two parts generator and discriminator. Iizuka et al. [7] utilize this network with some changes in the discriminator part to achieve better results for images that are both locally and globally consistent. In this work, two discriminators are used, in which the global discriminator checks that the image is whole consistent while the local one only looks at the region in which the missing block is centered. The advantage of this network is that it could generate blocks that do not appear in the image. So, this is good for pictures with highly specific structures like face [7].

### C. Text and Small Blank Removing Based Methods

This type of problem deals with finding text and removing it by using inpainting methods. Its application is in eliminating subtitles, logos, and stamps from images or video frames, somehow aesthetic retains. In [8], a combination of finding text and inpainting methods are used to design an automatic text removing system. In this case, inpainting methods are performed for both the structure and texture of masked images. Also, a simple morphological algorithm is used as the link of text detection and inpainting masked images. The proposed method in [9] has also contained two parts of text detection and de-occlusion of text regions. In this paper, as text detection, connected component labeling is performed. In the following, a set of selection/ rejection criteria is used to extract only text regions. Then the detected text region is inpainted by using a fast marching algorithm. In [10], for removing text regions, two steps are defined. In the first step, by morphological operation and connected component, the text region is detected and in the second step, by using a patching based method, the text is removed.

In some algorithms, more than one of the mentioned degradations are involved. As an example, in [11], the proposed method uses a fast convergent image inpainting method to deal with all of the previous degradation types. To achieve this goal, a combination of Richardson extrapolation and an improved BSCB inpainting model is used. In [12], irregular holes are considered as missed regions. The shape of these degradations is different, and therefore the proposed method should consider this point. For solving the ambiguity that exists in the missing areas, their proposed method uses an adversarial model that contains two stages. In the first stage of the proposed model, an edge generator is utilized to predict the edge of missed pixels. In the following, a completion network inpaints missed regions by using prior knowledge of edges, which is available from the previous stage. In [13] also, a different type of damages is addressed by using a patch- sparsity-based algorithm. In this paper, an Exemplar-based inpainting algorithm is combined with a facet model to achieve better results.

In this paper, a multi-scale method is proposed to provide local and global weights with lost pixels. In this method, one-dimensional interpolation in four directions is applied on different scales of the image, and in the end, the average of multi-scale paths is used as the final result. In the following, the proposed method is explained in section II, experimental results are shown in section III, and eventually, a concise conclusion is presented in section IV.

## II. PROPOSED METHOD

High-frequency changes in natural images rarely occur. Due to this, there is a local correlation between adjacent pixels. However, besides local information, the value of pixels also depends on the global information of the image. Considering both local and global features of images at the same time is an essential issue in image processing especially image inpainting. For this purpose, a multi-scale image inpainting method is proposed to combine both local and global data to predict missing pixels. The proposed method consists of different blocks, (i) selection of the number of scales, (ii) multi-scale formation, (iii) adaptive inpainting, (iv) up-sampling and (v)voting. The block diagram of the proposed method is shown in Fig. 1. In the following, these blocks would be explained in detail.

### A. Select Number of Scales

In image inpainting, two inputs are required, (i) the damaged image and (ii) a mask. As mentioned before, the size and the shape of missing pixels are different, and this is an essential issue in image inpainting proposed methods. In this paper, the width of the noise is the main parameter that identifies the number of scales of the input image. To deciding how to choose proper scales, two strategies have been studied. In the first strategy, sequential factors of two are selected, and in the second one, the input image size decreases with sequential integer scales. In the first strategy, also known as a pyramid method, the maximum number of scales could be calculated with Eq. 1. In this equation, parameter 'w' is the width of noise, and the parameter $'m'$ is the maximum number of scales. This equation defines the maximum number of scales, not the optimum number. In most of the situation, the best scale is less than the maximum value. If the quantity of scales exceeds this, there is no damaged pixel in the mask, and the interpolation is not meaningful.

$$m = \lfloor \log_2 w \rfloor \tag{1}$$

In the second strategy, the maximum number of scales could be calculated with Eq. 2. The parameters in this equation are the same as before.

$$m = w - 1 \tag{2}$$

In this paper, experimental results show that the second strategy has better performance. Therefore, this strategy is selected to evaluate the proposed method. It should be mentioned that using more scales leads to utilizing more global information and so blurriness of output image. Thus the maximum number of scales is not the right choice. For achieving a formula to find the right choice for the number of scales in each mask thickness, five masks with different width are used. According to experimental results, Eq. 1 is a good option for this block.

### B. Multi-Scale Formation

An image in different scales has different levels of features, similar to features extracted by various layers of a neural network. All information details are in the input image, but when we downscale the size of the image, only the low-frequency components remain in the image. In other words, only the high-level information remains in the downsampled image. The lower the scale, the more global information remains in image. In this phase of the proposed method, after identifying the number of the scales, the downsampled versions of the input image are created. There are different approaches to resize an image. For this application, we need an approach that explicitly uses the value of pixels without changing them. To achieve this goal, "*the nearest neighbor approach*" is selected. In this algorithm, the value of the closest pixel is used for the considered downsampled position. This phase of the algorithm is applied to both inputs, namely the damaged image and the mask. The scale versions of these inputs are sent to the next phase of the algorithms.

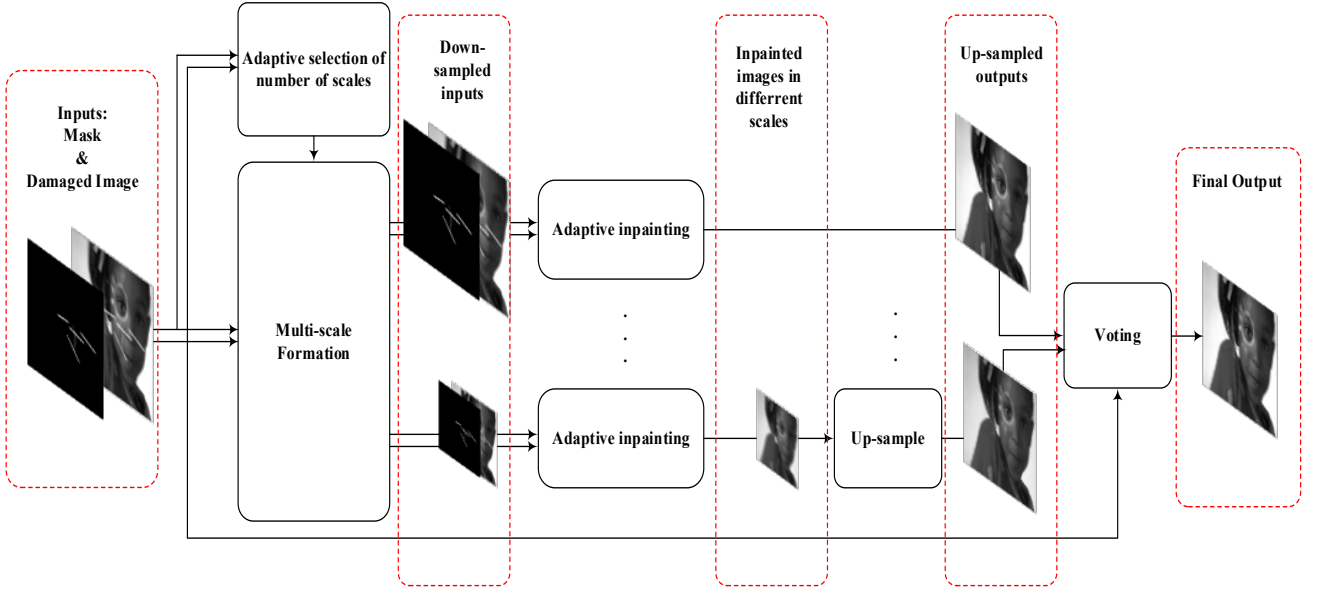

**Fig. 1. Block diagram of the proposed method**

### C. Adaptive Inpainting

According to the previous work presented in [14], this block consists of three sub-blocks, (i)pixel selection, (ii) directional interpolation, and (iii) fusion. In the first step, by using the mask, the lost pixels are separated from the whole pixels for processing. Due to the different sizes of the input image, the lost pixel for each scale is different. Hence, all of the steps in this phase are separately performed for each scale. In the second step, for each lost pixels, one-directional interpolation is applied to predict the value of it. For detecting the best neighbors to determine the values of lost pixels, the prediction is applied in four directions, horizontal, vertical, 45 and 135 degrees. Because of the importance of edge in various directions, this step also checks the existence of horizontal and vertical edges. In the fusion step, the information of edges is checked, if the location of the pixel is on the edge, then the predicted value of that direction is used. In other situations, by checking the difference of four predicted values, the invalid value is omitted, and the average of remaining values is calculated for the selected missing pixel. In the previous work, a fixed window for checking edges is considered, but in our current work, besides using global features as the advantage of multi-scaling, it also employs different window sizes for checking the edges. This strategy results in more accurate edge detection of our method as compared to previous methods. After running this block, for each scale, one output is produced, and these results are as the input of the next block.

### D. Up-sampling

From the previous block, outputs with different scales are generated. Because the size of each output is different, we need an up-sampled block before merging the outputs. Similar to the downsampling step, there are many methods for up-sampling. In this block, the low-resolution output should be converted to a high resolution one, so it needs an accurate selection for the resizing process. The approach chosen for this purpose is the bi-cubic spline. In this algorithm, values of pixels, which are before and after the selected pixel, are used to interpolate the median pixel.

### E. Voting

We have as many outputs from the up-sampling step as there are scales. These outputs will be used to generate the final result. For creating each output in the adaptive inpainting block, we use a restrictive approach, based on edge and directional information. Hence, the output of each scale is valid, and there is no misleading value between them. In other words, each of multi-scale output compensates for the shortcomings of other scales based on the local and global views. The experimental results also emphasize these multiscale views. In the next section, these results would be shown. Due to this, for final output, the weighted average of them is utilized. The weights of each value depend on the number of scales and should be set experimentally. In this paper, for detecting the best number of scales, equal weight is considered.

## III. EXPERIMENTAL RESULTS

As mentioned before, the influence of multi-scale for using local and general information of background pixels simultaneously is the motivation of predicting the missed value of masked regions. In the following, at first, the implementation details are noticed. The effect of the number of scale layers is studied in the second step, and finally, comparison with other methods is performed.

### A. Implementation Details

The hardware used for implementing all the codes is a core i7 CPU @2.5GHz with 8GB RAM. Also, MATLAB 2018a is used as the implementation software. Fifteen images of the Kodak dataset are utilized for evaluation and comparison with other methods. The reported metrics are average of the metrics on these images. At first, these images are masked. The masks used in this step are a simulation of scratches. As mentioned before, the location of lost pixels is important for the result of methods. So, simulated scratches are lines with random location, direction, and length. The thickness of these lines is a parameter that should be considered in specifying the number of scale layers. In the following, the effect of line thickness on the number of utilized scale layers is discussed.

### B. Mask Thickness Effect

In image inpainting, there are two important factors that influence the performance of reconstructed images, (i) location of scratches and (ii) thickness of scratches. The location of scratches could be in texture or flat regions of images. As mask lines become thicker, more general information about lost pixels should be used to reconstruct them properly. In these situations, selecting local and global neighbors have an important effect. In the proposed method, by using different scales of an image, both local and global information are used simultaneously. This improves the results of adaptive inpainting, which proposed in previous works [14].

As mentioned before, two strategies are tested for selecting the number of scales. The metric used for the comparison of these two strategies is Peak Signal to Noise Ratio (PSNR). For this purpose, every image in the Kodak dataset has been masked with five random masks that are made by lines with a thickness of five pixels. For both strategies, the average PSNR of reconstructed images has been reported in Fig. 2. The results of the first strategy are shown in red, and the second has been shown in blue. An equal number of scales used in each approach, preference of the first strategy is more than the second one. Hence, in the following steps, the first strategy is used.

It also mentioned that using more scales in the proposed multi-scale approach doesn't necessarily lead to better results. More scales result in smaller versions of input images, and hence, more global information is extracted. The scale-down of the image leads to partial disappearance of edges and textures of reconstructed images. To avoid this problem, a proper number of scales, based on the thickness of missing lines, should be selected. For this purpose, the algorithm is applied to different scales. The results are shown in Fig. 3. In this figure, the effect of the number of scales on the reconstructed output image is shown with PSNR. The bullets represent maximum PSNR and its corresponding number of scales for each mask with determined line thickness. Eq. 1 describes the relationship between mask thickness and the proper number of scales to be utilized.

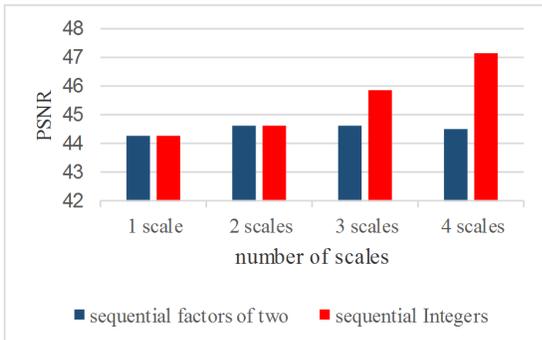

Fig. 2. Comparison of two strategies used for selection of scales.

### C. Comparison with Other Methods

To compare the proposed method with similar methods, which could be used in the same application, methods in [4], [5], and [10] have been used. Evaluation of these methods has been performed by using PSNR, Structural Similarity (SSIM), and run time metrics. In [5] a method is presented that finds a proper patch for lost pixels, and hence it would search the whole image that is a time-consuming process. Furthermore, as this method is patch-based, pixel-wise details are considered less, and so, its PSNR and SSIM are lower than the proposed method of this work. In [4], the methods used neighbor pixels and so it has better performance in PSNR and SSIM metrics. But in comparison with [14], adaptive selection of a number of lost pixels' neighbors in [14] leads to more informative content and as a result better performance. However, the draft of [14] is weak in utilizing more general information for inpainting lost pixels, and so, this study handles this defect by using multi-scale inpainting information. The results of the mentioned methods are compared with each other in Table 1. As is seen in Table 1 our approach has better performance in PSNR with a little decrease in SSIM. It also outperforms from [5], [4] in run time.

The output of our proposed method and the methods which are compared with them are shown in Fig. 4, as it could be seen by using global information our algorithm can predict missed pixels better than others. Although the proposed method on average outperforms other mentioned methods, definitely there is imperfection and constraint. As the proposed method in this work and the presented one in [14] uses just horizontal and vertical edge information, if scratches occur in regions that contain edges with other directions, proposed methods miss edge information and so they don't perform well. In Fig. 5. results of methods mentioned in [4], [14] and our approach are shown. The test image in this figure has lots of edges in different directions and so the results of the proposed methods are not satisfactory. As bi-cubic spline is used in [4], it could present better prediction in this case.

### IV. CONCLUSION

Image inpainting is a technique for image enhancement. As mentioned before, different types of damage could corrupt the image. In this work, the focus is on scratches and crease types, which their shapes are lines with different directions. In this paper, based on local and global information, a multi-scale image inpainting method is proposed. Our approach consists of different blocks, including (i) selection of the number of scales,

TABLE I.    The Results of PSNR, SSIM AND TIMES for Our Method and Methods in [4]-[5]-[10]

| The methods | Metrics | | |
|---|---|---|---|
| | *PSNR* | *SSIM* | *Times (sec)* |
| **Motmaen [4]** | 44.14 | **0.9965** | 21.1251 |
| **RIBBONS [5]** | 31.33 | 0.9776 | 2.4702 |
| **our previous method [10]** | 44.57 | 0.9944 | **1.89** |
| **Ours** | **45.41** | 0.9949 | 4.4 |

(ii)

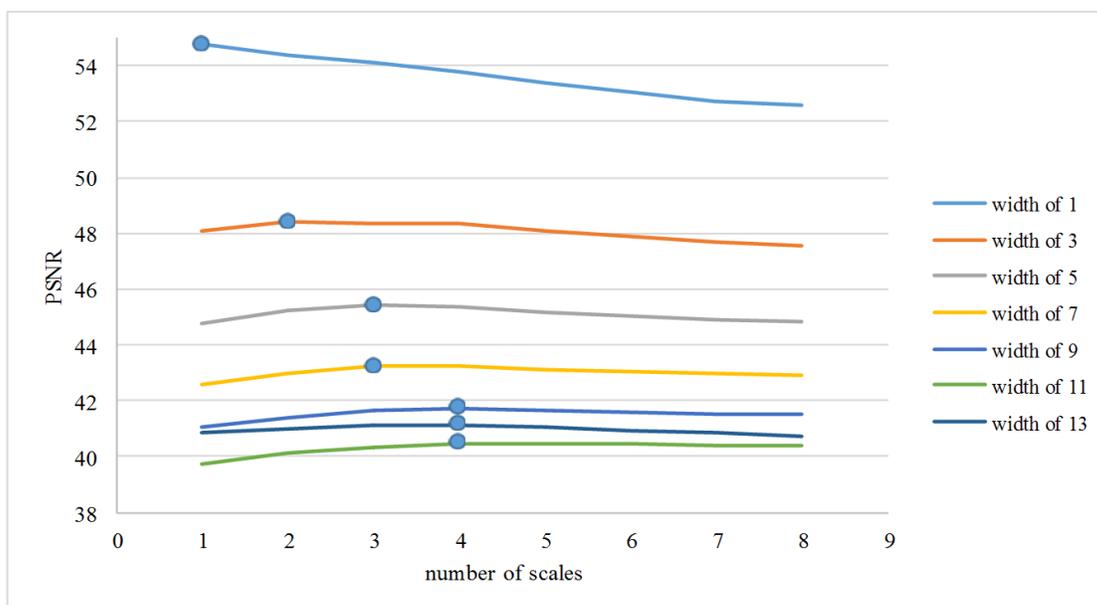

**Fig. 3. Proper number of scales according to best PSNR for different mask thickness**

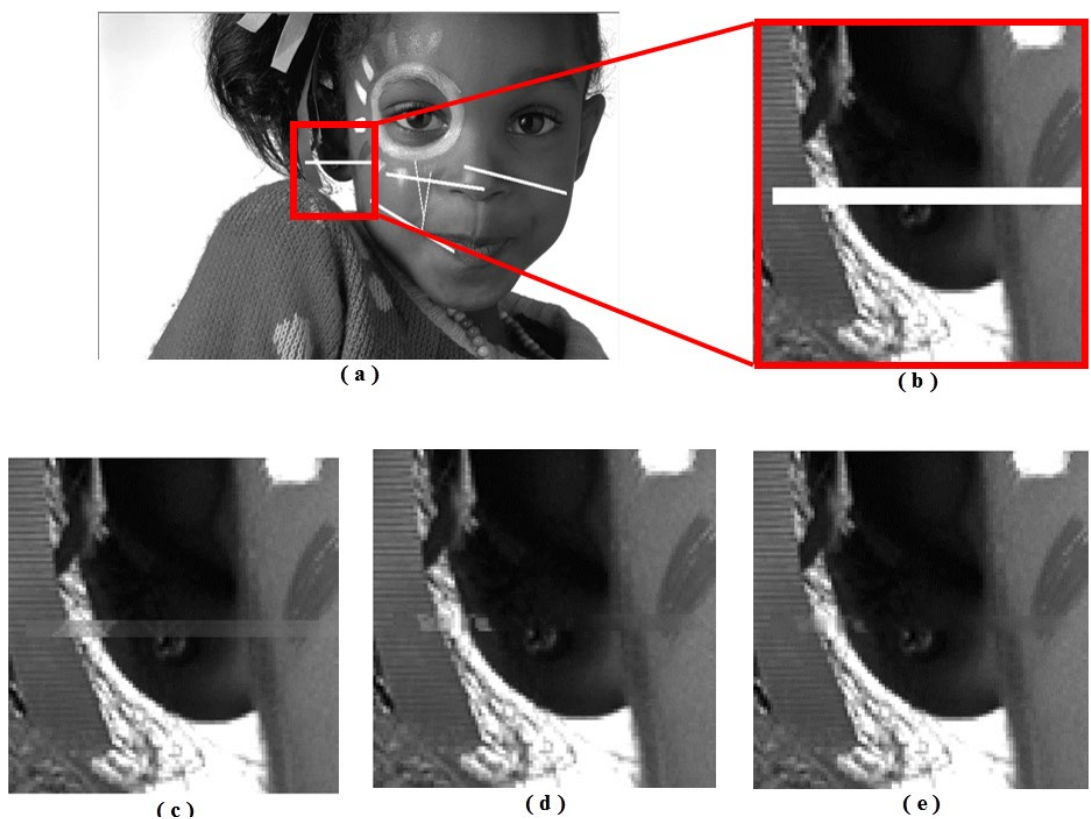

**Fig. 4. Our method outperforms methods of [4] and [14]. (a) Masked image, (b) zoomed image for better presentation, (c) results of method [4], (d) result of [14] and (e) results of the proposed method.**

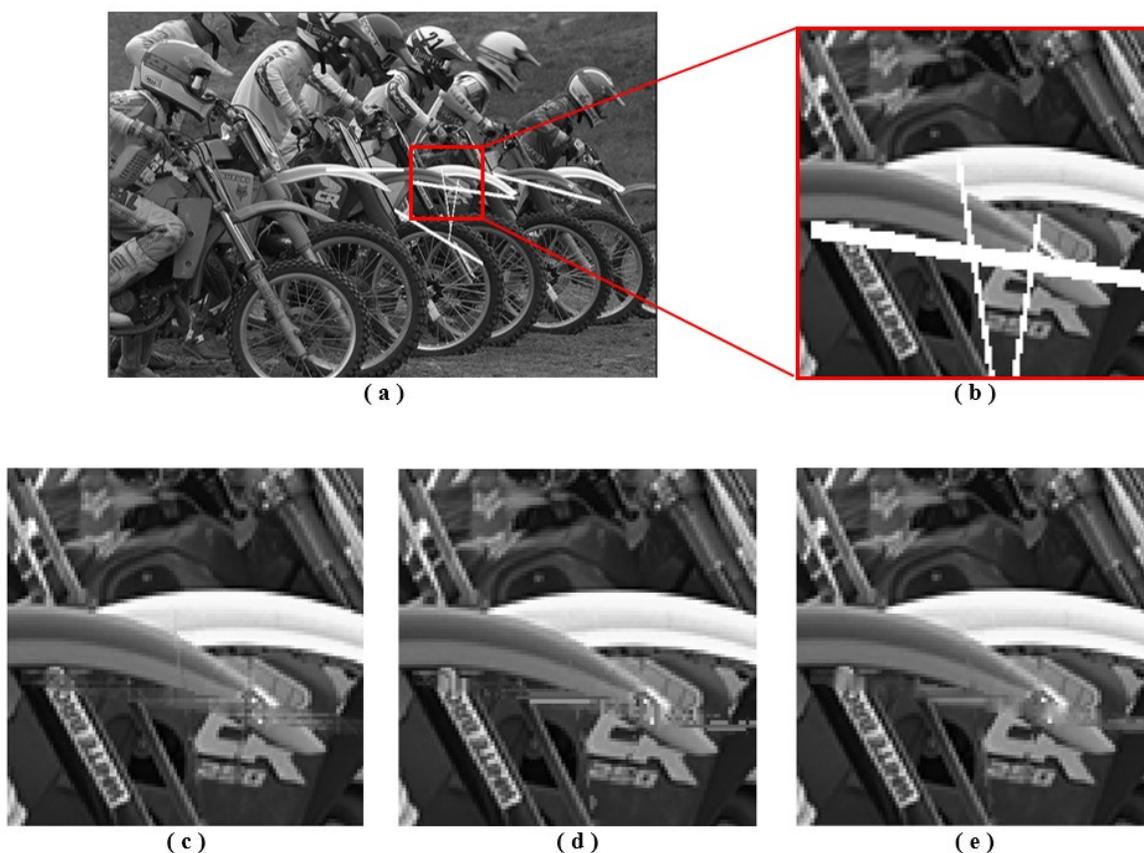

**Fig. 5. Fails of our method compared to method of [4]. (a) Masked image, (b) zoomed image for better presentation, (c) results of method in [4], (d) results of method in [14], and (e) results of the proposed method.**

multi-scale image formation, (iii) adaptive inpainting, (iv) up-sampling, and (v) the voting phase. By adding multi-scale to our previous work, two advantages are achieved. Firstly, by using both local and global features, we found missing pixels in flat and textured regions. Secondly, we detected edges very accurately by employing different window sizes. The experimental results show that our method caused improvements. By using PSNR for evaluation of our approach, we showed that on average, there is 1 dB improvement compared to similar works.